# Technical Report for Ego4D Long Term Action Anticipation Challenge 2023

Tatsuya Ishibashi, Kosuke Ono, Noriyuki Kugo and Yuji Sato

Panasonic Connect Co., Ltd.
{ishibashi.tatsuya001, ono.kosuke, kugou.noriyuki, sato.yuji}@jp.panasonic.com

## Abstract

*In this report, we describe the technical details of our approach for the Ego4D Long-Term Action Anticipation Challenge 2023. The aim of this task is to predict a sequence of future actions that will take place at an arbitrary time or later, given an input video. To accomplish this task, we introduce three improvements to the baseline model, which consists of an encoder that generates clip-level features from the video, an aggregator that integrates multiple clip-level features, and a decoder that outputs Z future actions. 1) Model ensemble of SlowFast and SlowFast-CLIP; 2) Label smoothing to relax order constraints for future actions; 3) Constraining the prediction of the action class (verb, noun) based on word co-occurrence. Our method outperformed the baseline performance and recorded as second place solution on the public leaderboard.*

## 1. Introduction

Ego4D [1] is a diverse and large first-person video dataset, and long-term action anticipation is one of the key tasks in Ego4D. The aim of this task is to realize a model that predicts $Z$ future actions for an input video of arbitrary length.

Our contributions are summarized below:

1. We introduce three improvements to the baseline model, which consists of an encoder that generates clip-level features from the video, an aggregator that integrates multiple clip-level features, and a decoder that outputs $Z$ future actions. 1) Model ensemble of SlowFast and SlowFast-CLIP; 2) Label smoothing to relax order constraints for future actions; 3) Constraining the prediction of the action class (verb, noun) based on word co-occurrence.

2. On the public leaderboard, our proposed model improves by 0.0331, 0.0574, and 0.0320 points over the baseline model prediction for verb, noun, and action, respectively.

## 2. Our Approach

### 2.1. Overall Architecture

Our network architecture is shown in Figure 1. In this architecture, we first input the video clips to the Video Encoder, which extracts features from each clip. Then, the Feature Aggregator merges the extracted features. Next, the Multi-Head Decoder takes these features as input and generates output logits for each of the $Z$ heads associated with nouns and verbs. After that, we perform an ensemble by combining decoder outputs using weighted sum Moreover, we refine the output results using statistical measures regarding the co-occurrence of verb and noun labels calculated from training and validation data to obtain the final prediction results. The following sections explain Encoder-Decoder Architecture, Model Ensemble, and Refinement Module.

### 2.2. Encoder-Decoder Architecture

**Video Encoder**
In our approach, we used two types of video encoders: SlowFast [2], which extracts temporal features from $I$ video clips, and CLIP [3], which extracts features related to relationships between objects and actions. In addition, it is reported in the baseline paper [1] that increasing the number of input clip videos to be encoded and observing more past videos allows predictions to consider the context of more past videos, thereby increasing accuracy. In our approach, we also added a model with increased input clip videos.

**Feature Aggregator**
The clip-level features generated by the encoders are then integrated by the subsequent aggregator. In our approach, we used Concat and Transformer, two methods that are introduced in the baseline method.

**Multi-Head Decoder**
Finally, the features of the entire observed video generated by the aggregator are passed to the decoder, which outputs sequences of actions at future time steps. Again, we used the Multi-Head technique introduced in the baseline method, which predicts $Z$ actions by independent heads.

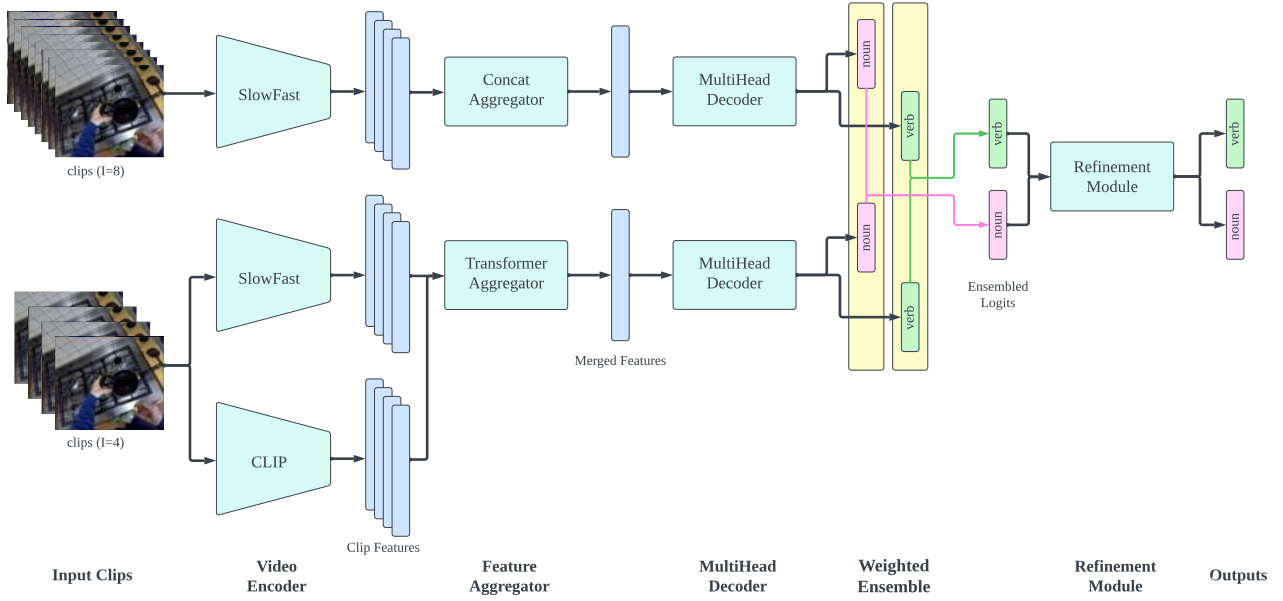

**Figure 1** Overall architecture of our approach.

### 2.3. Weighted Ensemble

The two types of encoders used in Video Encoder (SlowFast and SlowFast-CLIP) have different characteristics: SlowFast is more accurate in predicting verbs by capturing temporal features, while CLIP is more accurate in predicting nouns by focusing on objects. Given this difference in characteristics, our approach employs an ensemble approach in which the two Encoders complement each other's inference of verbs and nouns, with the expectation that this will improve the prediction accuracy of the action. Specifically, the logits output from each model are combined by a weighted sum.

$$\boldsymbol{Logits} = \boldsymbol{\alpha Logits_{SF}} + \boldsymbol{\beta Logits_{SF-CLIP}} \quad (1)$$

### 2.4. Refinement Module

This study proposes a method for output refinement to consider contextual relationships and improve output consistency. The baseline method predicts verbs and nouns separately, resulting in a lack of consistency between them. Since the predicted classes are randomly selected based on the prediction probability distribution at each time step, there is no consistency within the sequential prediction patterns. As mentioned in [1], the co-occurrence of words, such as normalized pointwise mutual information (NPMI)

[4], is considered important for long-term predictions. We have arranged the equation presented in [5] and introduced an indicator that considers the relationships between consecutive verbs and nouns. Given a time step $z$ with the class label $x_n$, the formula is as shown below:

$$f(x_{z-1}, x_z) = \ln\left(\frac{p(x_z|x_{z-1})}{p(x_{n-1})p(x_n)}\right) / -\ln(p(x_z|x_{z-1})) \quad (2)$$

Furthermore, we compute the probability $g$ of a verb $V$ occurring simultaneously with a given noun $N$ as follows:

$$g(V_z, N_z) = p(V_z|N_z) \quad (3)$$

We computed these statistics based on the training and validation sets. Finally, the predicted probability of verb $v$ and nouns $n$, $P_v^z$ and $P_n^z$, at time step $z$ are refined as shown in Equations (4) and (5) for each sequential prediction pattern.

$$\hat{P}_n^z = P_n^z \circ ReLU\big(f_{noun}(N_{Z-1}, n)\big) \quad (4)$$
$$\hat{P}_v^z = P_v^z \circ ReLU\big(f_{verb}(V_{Z-1}, v)\big) \circ g(v, N_z) \quad (5)$$

Moreover, we adopted a strategic approach to maintain consistency within the predicted patterns. For one of the predicted patterns, we selected the class with the highest prediction probability without performing output refinements, for another pattern, we selected the class with the highest prediction probability after output refinements. For the other patterns, we randomly selected based on the refined prediction probability distribution.

| | Encoder | Aggregator | $I$ | Label smoothing |
|---|---|---|---|---|
| A | SlowFast | Concat | 8 | |
| B | SlowFast+CLIP | Transformer | 4 | ✓ |

**Table 1** Parameter configurations used in the training of individual models. Baseline settings were adopted for all other parameters.

| Dataset | Model | Verb | Noun | Action |
|---|---|---|---|---|
| Validation | Baseline | 0.7039 | 0.6861 | 0.9173 |
| | Ours | **0.6702** | **0.6291** | **0.8753** |
| Test | Baseline | 0.7169 | 0.7359 | 0.9253 |
| | Ouers | **0.6838** | **0.6785** | **0.8933** |

**Table 2** Comparison of the baseline and proposed approach for validation and test data. The baseline consists of SlowFast encoder and Transformer aggregator. The scores on the test data are cited from the leaderboard.

| Method | Verb | Noun | Action |
|---|---|---|---|
| model A | 0.7053 | 0.7058 | 0.9232 |
| model B | 0.7046 | 0.6717 | 0.9139 |
| A+B | 0.6948 | 0.6563 | 0.9079 |
| A+B+refinement | **0.6618** | **0.6266** | **0.8762** |

**Table 3** Comparison of individual models and the ensemble model, as well as the results obtained when incorporating the refinement module.

| Method | Verb | Noun | Action |
|---|---|---|---|
| model B (w/o label smoothing) | 0.7068 | 0.6935 | 0.9189 |
| model B (w/ label smoothing) | **0.7046** | **0.6717** | **0.9139** |

**Table 4** Comparison on label smoothing

## 2.5. Label Smoothing

In the baseline method, the loss for each predicted time step was calculated as the cross-entropy between the one-hot ground truth labels and the predicted probabilities for verbs/nouns. In this case, a huge penalty is applied even if the step is off by one. However, the long-term action anticipation task is challenging to predict the order accurately. Therefore, we adopt a less stringent learning approach regarding order errors by using smoothed labels instead of one-hot ground truth values. Given the one-hot ground truth value represented as $y_z$, the smoothed label $y'_z$ is expressed as follows:

$$y'_z = \frac{y_z + \frac{1}{Z}\sum_{t=1}^{Z} y_t}{2}$$

## 3. Experiments

### 3.1. Implementation Details

For settings not explicitly mentioned, we follow the approach outlined in baseline method [1]. We trained two models as the foundation for our ensemble approach, with their respective training configurations detailed in Table 1. We used a pretrained checkpoint of SlowFast encoder model provided for long-term action anticipation task. In Model A, we employed the SlowFast encoder and a Concat aggregator, with the number of input clips set to 8. In Model B, we used two encoders, SlowFast and CLIP, along with a Transformer aggregator, setting the number of input clips to 4. Additionally, during the training process, we incorporated label smoothing as described in Section 2.5. For each training process, we used 4 NVIDIA Tesla V100 GPUs, utilizing a batch size of 32, a learning rate 0.0001, and spanning 50 epochs.

### 3.2. Main Results

In Table 2, we show a comparison between the baseline and our proposed approach on the validation and test set. According to the official guidelines, the number of actions to predict $Z$ was set at 20, while the number of output patterns $K$ was set at 5. We conducted a thorough search for the parameters of the ensemble weights, aiming to improve the accuracy on the validation data. As a result, we determined the optimal parameters to be $\alpha = 0.6$ and $\beta = 1.4$, respectively.

To investigate the effect the ensemble approach, we performed a comparative analysis between the individual models and the ensemble model, shown in Table 3. In this evaluation, the ensemble weights were set to $\alpha = 0.5$ and $\beta = 0.5$. By ensembling the two models, improvements in edit distance values were observed: approximately 0.010 for verbs, 0.015 for nouns, and 0.006 for actions. In the case of actions, the edit distance is calculated considering the predictions of verb and noun pairs. The smaller improvement in actions compared to individual results for verbs and nouns can be attributed to the potential loss of consistency resulting from the ensemble of multiple models. Moreover, output refinement improved 0.032 point for actions, representing the most substantial improvement in our method. This suggests that considering the contextual relationship with past actions has a positive impact on the performance. The improvement in action was comparable to those of verb and noun performance, indicating the effectiveness of considering the co-occurrence of verbs and nouns.

Table 4 shows the effect of label smoothing, which was applied only during the training of Model B. The results of both verb and noun are improved, with a 0.05 point enhancement in action prediction. This can be attributed to the suppression of penalties related to sequence order misalignments, similar to the evaluation metric of edit distance.

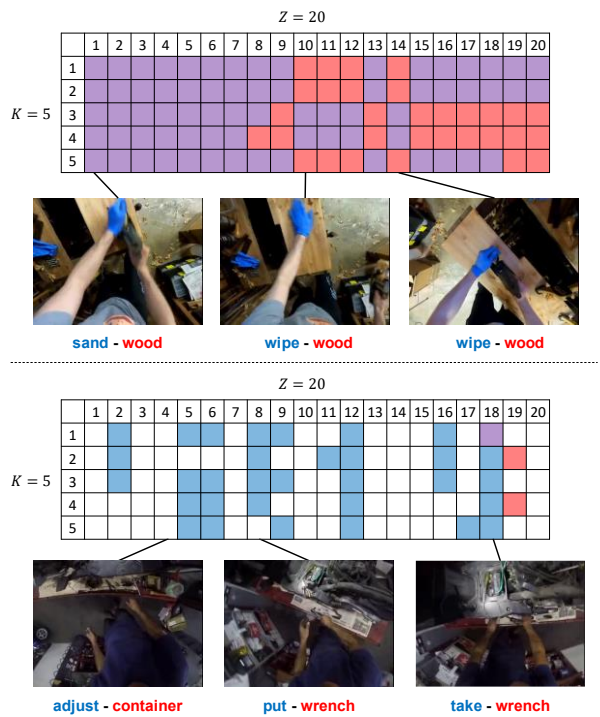

**Figure 2 Positive and negative cases. The grids show prediction patterns in rows and prediction target steps in columns. Blue cells represent correctly predicted only verbs, red cells represent correctly predicted only nouns, and purple cells represent where both verbs and nouns are correct. The texts bellow the images represent ground-truth labels (Blue: verb, Red: noun)**

**3.3.** Examples of Positive and Negative Results

Figure 2 shows the examples of that are correctly and incorrectly predicted by our approach. In accurate cases, fewer prominent objects are present in the image, and the number of action patterns is limited. In this case, there are only two types of actions: "sand wood" and "wipe wood." Furthermore, the proposed approach has become more likely to make consecutive predictions within the same pattern due to considering temporal relationships in the output refinement module.

On the other hand, common characteristics of failed cases include many objects with which people can easily interact, significant changes in the field of view due to the movement of subjects, and high complexity of actions meaning that there are many possible actions that can be taken towards a single object.

## 4. Conclusion

In this report, we introduce three improvements over the baseline in the long-term action anticipation task for first-person videos. Results on the validation and test set show that the proposed method can achieve excellent performance.